\documentclass[10pt,twocolumn,letterpaper]{article}
\pdfoutput=1
\usepackage[pagenumbers]{iccv}      

\usepackage{appendix}
\usepackage{multirow}
\usepackage{marvosym} 
\usepackage[most]{tcolorbox}
\usepackage{listings}
\usepackage{graphicx}
\usepackage{amsmath}
\usepackage{amssymb}
\usepackage{paralist}
\usepackage{array}

\definecolor{lightgray}{gray}{0.95}
\definecolor{deepblue}{RGB}{70,130,180}
\definecolor{deepgray}{RGB}{119,136,153}
\lstdefinestyle{prompt}{
    basicstyle=\ttfamily\fontsize{7pt}{8pt}\selectfont,
    frame=none,
    breaklines=true,
    backgroundcolor=\color{lightgray},
    breakatwhitespace=true,
    breakindent=0pt,
    escapeinside={(*@}{@*)},
    numbers=none,
    numbersep=5pt,
    xleftmargin=5pt,
    aboveskip=2pt,
    belowskip=2pt,
}
\definecolor{iccvblue}{rgb}{0.21,0.49,0.74}
\definecolor{color1}{RGB}{192,0,0}
\definecolor{color2}{RGB}{35,87,35}
\usepackage[pagebackref,breaklinks,colorlinks,allcolors=iccvblue]{hyperref}
\newcommand{\Paragraph}[1]{\noindent\textbf{#1}}

\newcommand{\sys}{Edit Transfer\xspace}

\def\eg{\textit{e.g.}\xspace}

\def\vs{\textit{vs.}\xspace}

\newcommand{\subsecref}[1]{Section~\ref{subsec:#1}}
\newcommand{\figref}[1]{Fig.~\ref{fig:#1}}
\newcommand{\tabref}[1]{Table~\ref{tab:#1}}

\title{\sys: Learning Image Editing via Vision In-Context Relations}

\author{
Lan Chen$^{1}$ \hspace{0.65cm} Qi Mao$^{1,\textsuperscript{\Letter}}$ \hspace{0.65cm} Yuchao Gu$^2$ \hspace{0.65cm} Mike Zheng Shou$^2$ \\
$^1$MIPG, Communication University of China \hspace{0.65cm} $^2$Show Lab, National University of Singapore \\
\url{https://cuc-mipg.github.io/EditTransfer.github.io}
}

\begin{document}

\begin{figure*}
\twocolumn[{
\renewcommand\twocolumn[1][]{#1}
\vspace{-4 mm}
\maketitle
\vspace{-3 mm}
    \centerline{\includegraphics[width=\textwidth]{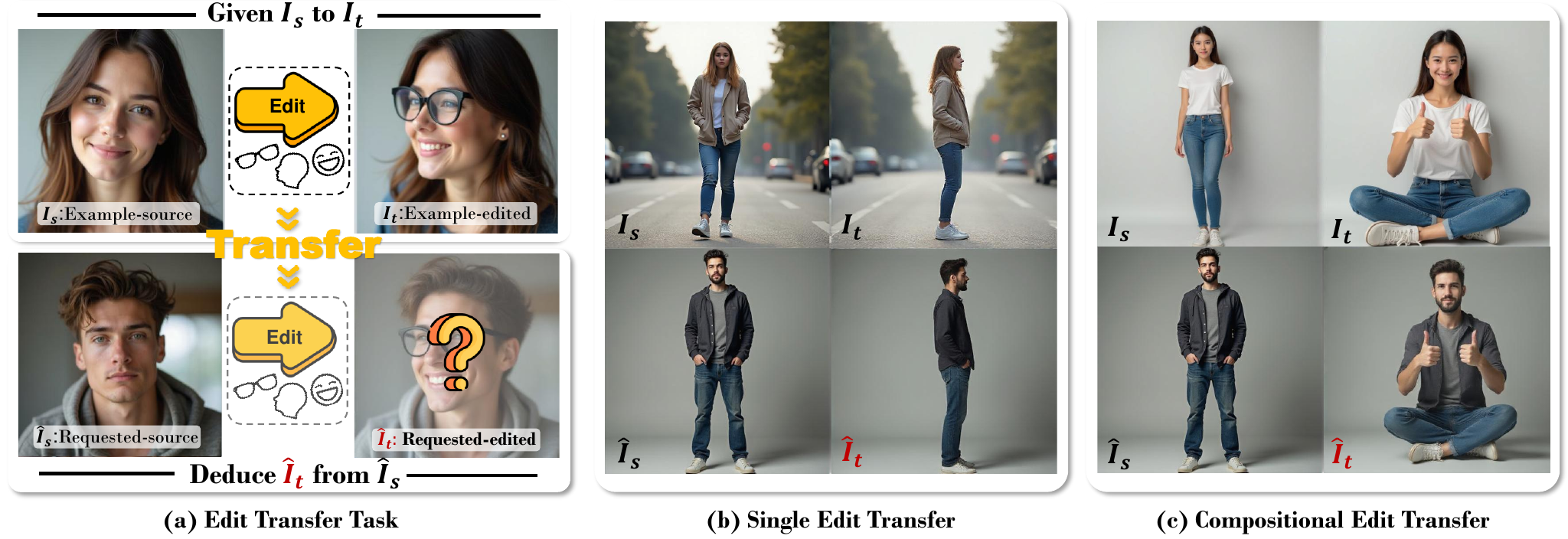}}
    \vspace{-2 mm}
    \caption{\textbf{Edit Transfer} aims to learn a transformation from a given source–target editing example, and apply the edit to a query image.
    Our framework can effectively transfer both (b) single and (c) compositional non-rigid edits via proposed visual relation in-context learning.
    }
    \label{fig:teaser}
    \vspace{3 mm}
}]
\end{figure*}
\let\thefootnote\relax\footnotetext{
\textsuperscript{\Letter} Corresponding Author}
\begin{abstract}
We introduce a new setting, \textbf{\sys}, where a model learns a transformation from just a single source–target example and applies it to a new query image.
While text-based methods excel at semantic manipulations through textual prompts, they often struggle with precise geometric details (\eg, poses and viewpoint changes).
Reference-based editing, on the other hand, typically focuses on style or appearance and fails at non-rigid transformations.
By explicitly learning the editing transformation from a source–target pair, \sys mitigates the limitations of both text-only and appearance-centric references.
Drawing inspiration from in-context learning in large language models, we propose a \textbf{visual relation in-context learning} paradigm, building upon a DiT-based text-to-image model. 
We arrange the edited example and the query image into a unified four-panel composite, then apply lightweight LoRA
fine-tuning to capture complex spatial transformations from minimal examples.
Despite using only $42$ training samples, \sys substantially outperforms state-of-the-art TIE and RIE methods on diverse non-rigid scenarios, demonstrating the effectiveness of few-shot visual relation learning.
\end{abstract}    
\section{Introduction}
\label{sec:intro}
Text-based image editing (TIE) has emerged as one of the most popular image generation scenarios, driven by advances in text-to-image (T2I) models~\cite{radford2021learning,ramesh2022hierarchical,rombach2022high,saharia2022photorealistic,esser2024scaling}. 
In TIE, a user provides a source image and a textual description of the desired edit, and the model generates the edited image accordingly.

\begin{figure}[!t]
    \includegraphics[width=\columnwidth]{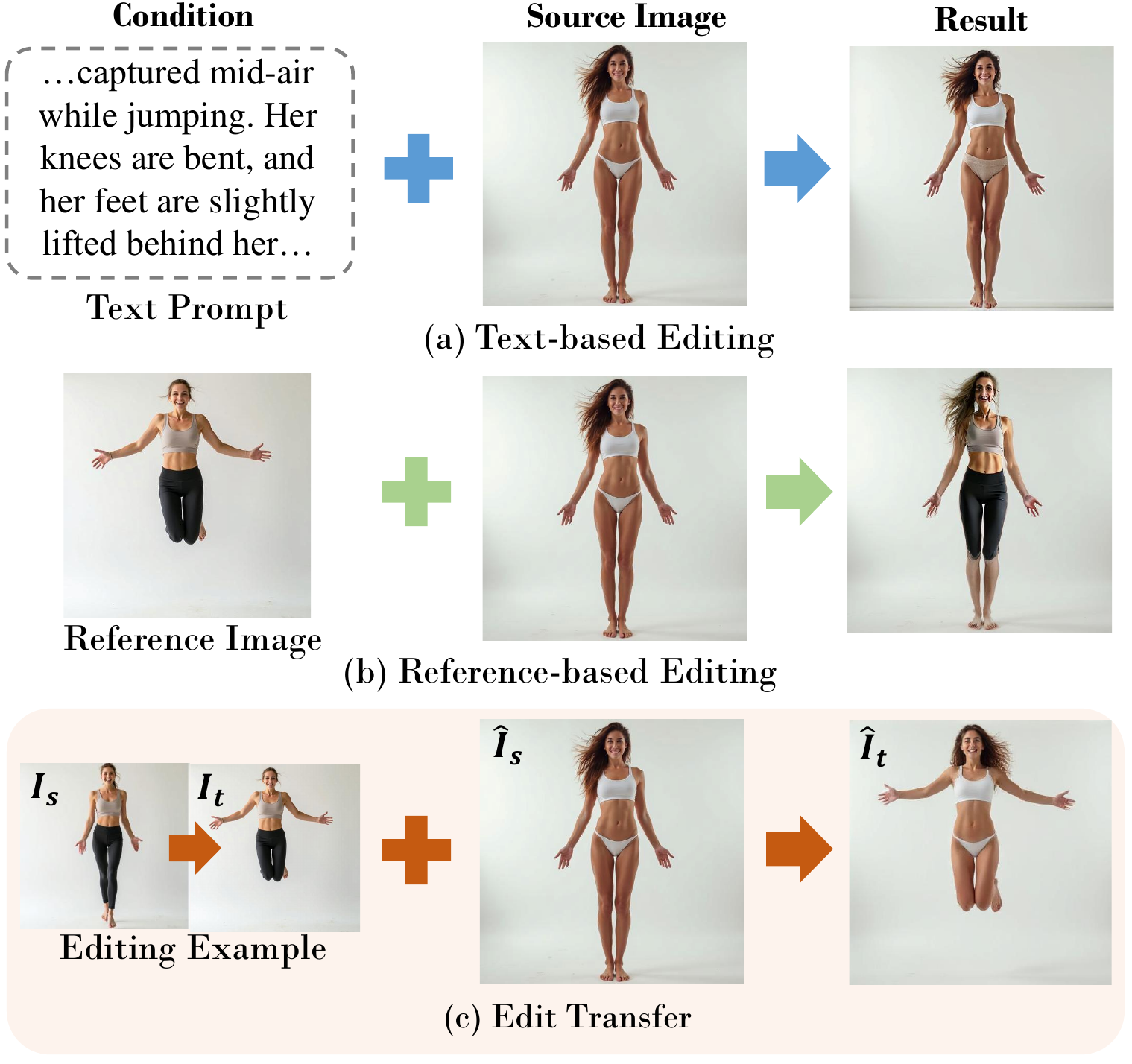}
    \vspace{-7 mm}
   \caption{\textbf{Comparisons with existing editing paradigms.}
(a) Existing TIE methods~\cite{hertz2022prompt,brooks2023instructpix2pix,cao_2023_masactrl,wang2024taming,avrahami2024stableflow,feng2024dit4editdiffusiontransformerimage} rely solely on text prompts to edit images, making them ineffective for complex non-rigid transformations that are difficult to describe accurately.  
(b) Existing RIE methods~\cite{Gatysstyle, Zhucyclegan, Alaluf2024transfer,zhou2025attentiondistillationunifiedapproach, YangPaintbyExample, ChenSpecRef, he2024freeedit, chen2024mimicbrush, biswas2025PIXELS,ChenAnyDoor} incorporate visual guidance via a reference image but primarily focus on appearance transfer, failing in non-rigid pose modifications.  
(c) In contrast, our proposed \sys learns and applies the transformation observed in editing examples to a query image, effectively handling intricate non-rigid edits.
    }
    \label{fig:compare}
    \vspace{-5 mm}
\end{figure}

However, not all editing requirements can be easily expressed through text. 
For instance, as illustrated in \figref{compare}(a), replicating precise arm or leg positions—such as the exact placement or jump height—can be challenging to describe purely through words.
This limitation arises because while text captures abstract semantics well, it does not convey the detailed spatial information needed for non-rigid, compositional transformations.

Motivated by these limitations, we ask a simple yet intriguing question: \emph{can we directly learn the transformation from a source image to its edited version, and then apply this transformation to a new query image?} 
Images naturally encode richer spatial cues than text, offering more accurate guidance for such complex manipulations.

Building on the idea of incorporating visual guidance,
various reference-based image editing (RIE) methods~\cite{Gatysstyle, Zhucyclegan, Alaluf2024transfer,zhou2025attentiondistillationunifiedapproach, YangPaintbyExample, ChenSpecRef, he2024freeedit, chen2024mimicbrush, biswas2025PIXELS,ChenAnyDoor} have been extensively explored.
However, existing approaches typically focus on transferring low-level properties such as style and texture (see \figref{compare}(b)), and often struggle with non-rigid spatial transformations, where more complex geometric manipulation is required.

In this paper, we propose a novel editing task, \textbf{\sys}, aiming to edit a query image by learning and applying the transformation observed in a given editing examples, as shown in \figref{teaser}(a).
Instead of relying on large-scale training with hundreds of thousands of samples, we pose a more challenging question: \emph{can such transformative editing be achieved using only a small set of paired images?}

Our inspiration draws from in-context learning~\cite{brown2020language} in large language models (LLMs), where a model can adapt to new tasks by observing a few example input-output pairs.
By analogy, we propose a \textbf{visual relation in-context learning} for \sys:
given a small number of ``source image $\rightarrow$ edited image'' pairs, the model is expected to transfer the demonstrated editing operation—whether it involves pose changes, facial adjustments, or other modifications—onto a new query image.

To achieve this, we build upon FLUX, a state-of-the-art text-to-image (T2I) model grounded in the DiT architecture~\cite{Peeblesdit}, which exhibits in-context learning capabilities in image generation~\cite{lhhuang2024iclora}.
We adapt FLUX for visual relationship learning by embedding both the example and query images into a single four-panel composite, merging them into one token sequence.
This unified representation allows the images to attend to one another through Multi-Modal Attention (MMA)~\cite{pan2020multi} in the DiT blocks. 
Additionally, we fine-tune the Low-Rank Adaptation (LoRA) layers 
 on a small dataset covering various editing types, strengthening the model’s capacity to capture and transfer complex visual transformations.
Remarkably, only $21$ editing types—each represented by two four-panel examples—suffice to achieve effective editing transfer (\figref{teaser}(b)).
Even more surprising, although each training sample reflects a single transformation, our model can seamlessly merge multiple editing operations present in an example pair (\figref{teaser}(c)), demonstrating impressive compositional abilities.

Our contributions can be summarized as follows:
\begin{compactitem}
\item We propose a new editing task, \sys, which learns the underlying transformation from a reference example and flexibly applies it to new images, enabling edits beyond the reach of TIE and RIE methods.

\item We introduce
a simple yet effective visual relation in-context learning paradigm, adapting a DiT-based T2I model with minimal data (just $42$ images in total), demonstrating that sophisticated editing behaviors can emerge from a handful of carefully designed examples.

\item Extensive experiments validate the effectiveness of our framework, particularly in handling complex spatial non-rigid transformations and composite editing tasks, outperforming state-of-the-art methods in various challenging scenarios.
\end{compactitem}

\section{Related Work}
\label{sec:related}
\begin{figure*}[!t]
\centering
    \includegraphics[width=\linewidth]{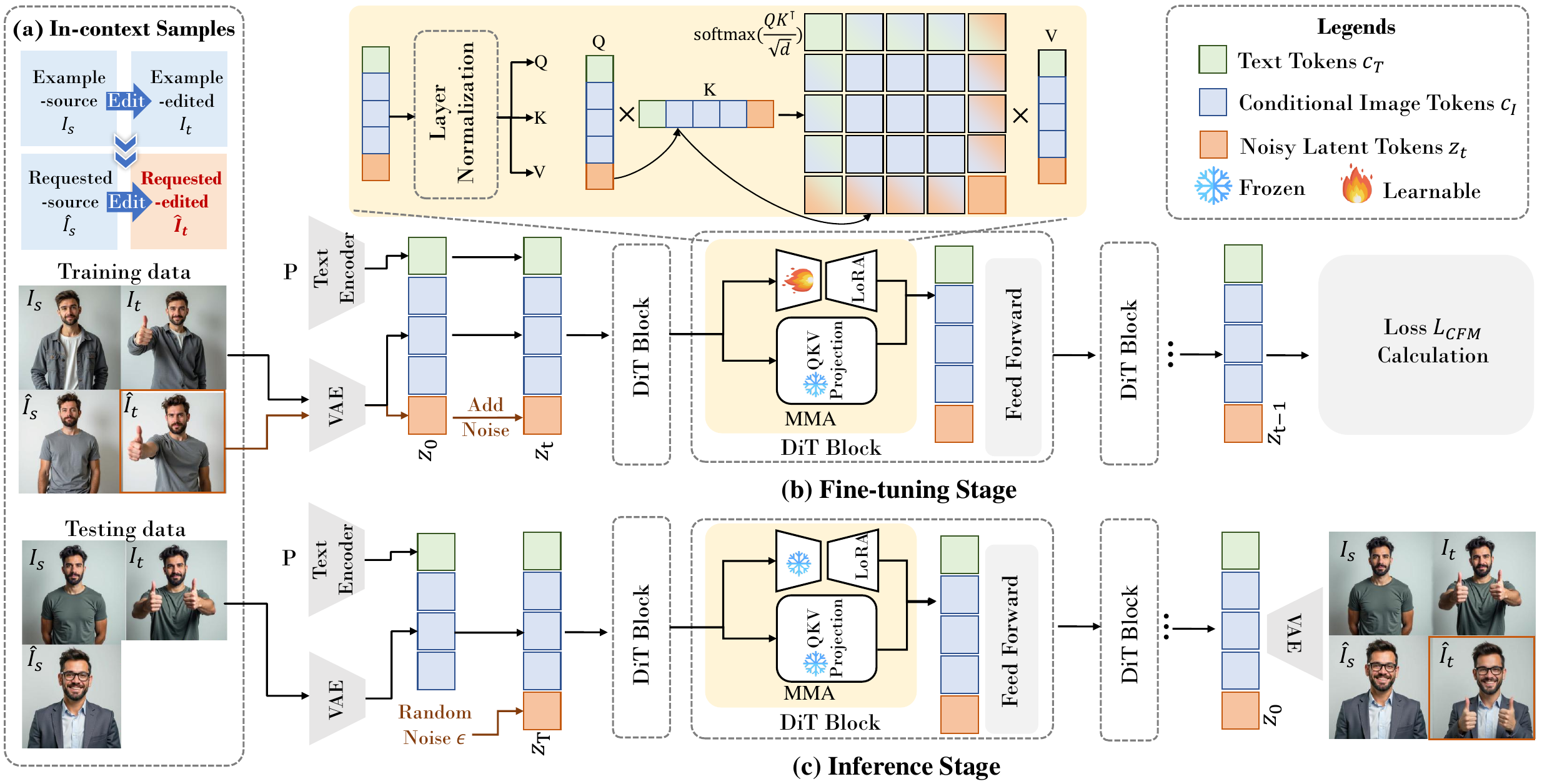}
    \vspace{-7 mm}
    \caption{\textbf{Visual relation in-context learning for \sys.}
    (a) We arrange in-context examples in a four-panel layout: 
the top row (an editing pair $(\mathcal{I}_s, \mathcal{I}_t)$) and the bottom row (the query pair  $(\mathcal{\hat{I}}_s, \mathcal{\hat{I}}_t)$).
Our goal is to 
 to learn the transformation from
$\mathcal{I}_s \to \mathcal{I}_t$, and apply it to the bottom-left image $\hat{\mathcal{I}}_s$, producing the target
$\hat{\mathcal{I}}_t$, in the bottom-right.          %
    (b) We fine-tune a lightweight LoRA in the MMA  to better capture visual relations.
Noise addition and removal are applied only to $z_t$, while the conditional tokens $c_T$ ( derived from $(\mathcal{I}_s, \mathcal{I}_t, \hat{\mathcal{I}}_s)$) remain noise-free.
            (c) Finally, we cast \sys as an image generation task by initializing the bottom-right latent token  $z_T$ with random noise and concatenating it with the clean tokens
$c_I$.
            Leveraging the enhanced in-context capability of the fine-tuned DiT blocks, the model generates
$\mathcal{I}_t$, effectively transferring the same edits from the top row to the bottom-left image.}
    \label{fig:framework}
    \vspace{-3 mm}
\end{figure*}
\Paragraph{Text-based image editing} aims to modify a given image based on user-provided text that describes the desired changes.
Methods such as P2P~\cite{hertz2022prompt} and Instruct-Pix2Pix~\cite{brooks2023instructpix2pix} successfully handle various edits, including changes in appearance, object replacement, and object addition.
However, they struggle with non-rigid edits, which require large geometric variations.
To address this limitation,
MasaCtrl~\cite{cao_2023_masactrl} provides a solution for simple non-rigid editing, such as lifting a dog's legs to a jumping pose.
Leveraging a DiT-based T2I model as backbone, recent works~\cite{wang2024taming,avrahami2024stableflow,feng2024dit4editdiffusiontransformerimage} achieve improved performance in handling non-rigid edits.
However, for edits involving complex spatial transformations, such as human poses that engage all limbs as illustrated in~\figref{compare}(a), text alone often fails to provide sufficient visual guidance, leading to unsuccessful edits.

\Paragraph{Reference-based image editing.}
Introducing visual guidance can compensate for the spatial details that are hard to describe with text alone, which has led to extensive research on RIE~\cite{Gatysstyle, Zhucyclegan, Alaluf2024transfer,zhou2025attentiondistillationunifiedapproach, YangPaintbyExample, ChenSpecRef, he2024freeedit, chen2024mimicbrush, biswas2025PIXELS,ChenAnyDoor}  using additional reference images.
Early style transfer methods~\cite{Gatysstyle, Alaluf2024transfer} focus on transferring the global artistic style of the reference image to the target.
Other approaches~\cite{Zhucyclegan,zhou2025attentiondistillationunifiedapproach} achieve appearance transfer by matching semantically aligned regions, for example, transferring textures between corresponding objects. 
To enable finer control, recent methods~\cite{ChenAnyDoor, YangPaintbyExample, ChenSpecRef, he2024freeedit, chen2024mimicbrush, biswas2025PIXELS} aim to transfer the appearance of specific local regions. 
Despite these advances, such techniques remain confined to low-level appearance transfer and often struggle to capture intricate spatial relationships, such as pose transformations (see~\figref{compare}(b)). 
 In contrast, we introduce \sys, a novel paradigm that transfers the visual relationship exemplified by an editing pair to a new query image, enabling complex spatial transformations beyond the reach of conventional RIE methods, as demonstrated in~\figref{compare}(c).

\Paragraph{Vision In-context learning.} 
In-context learning—the process of inferring patterns from provided input-output pairs—has been widely adopted in LLMs ~\cite{brown2020language}.
In the visual domain, in-context learning is initially explored for understanding tasks~\cite{Wangspeak, Barvisual, Zhanggood, bai2024sequential}.
For instance, \cite{bai2024sequential} demonstrates that in-context learning is effective for downstream representation tasks, such as segmentation.
In the field of visual generation, leveraging DiT-based text-to-image (T2I) models, IC-LoRA~\cite{lhhuang2024iclora} reveals that these models naturally exhibit in-context generation abilities.
It focuses on generating a set of sub-images with a consistent style or synthesizing images conditioned on a reference image.
In contrast, our work aims to address the proposed edit transfer task by extending the visual in-context learning paradigm, enabling the model to infer and apply transformations demonstrated in an editing exemplar to query images.
\section{Methodology}
\label{sec:method}
\begin{figure*}[t]
    \includegraphics[width=\linewidth]{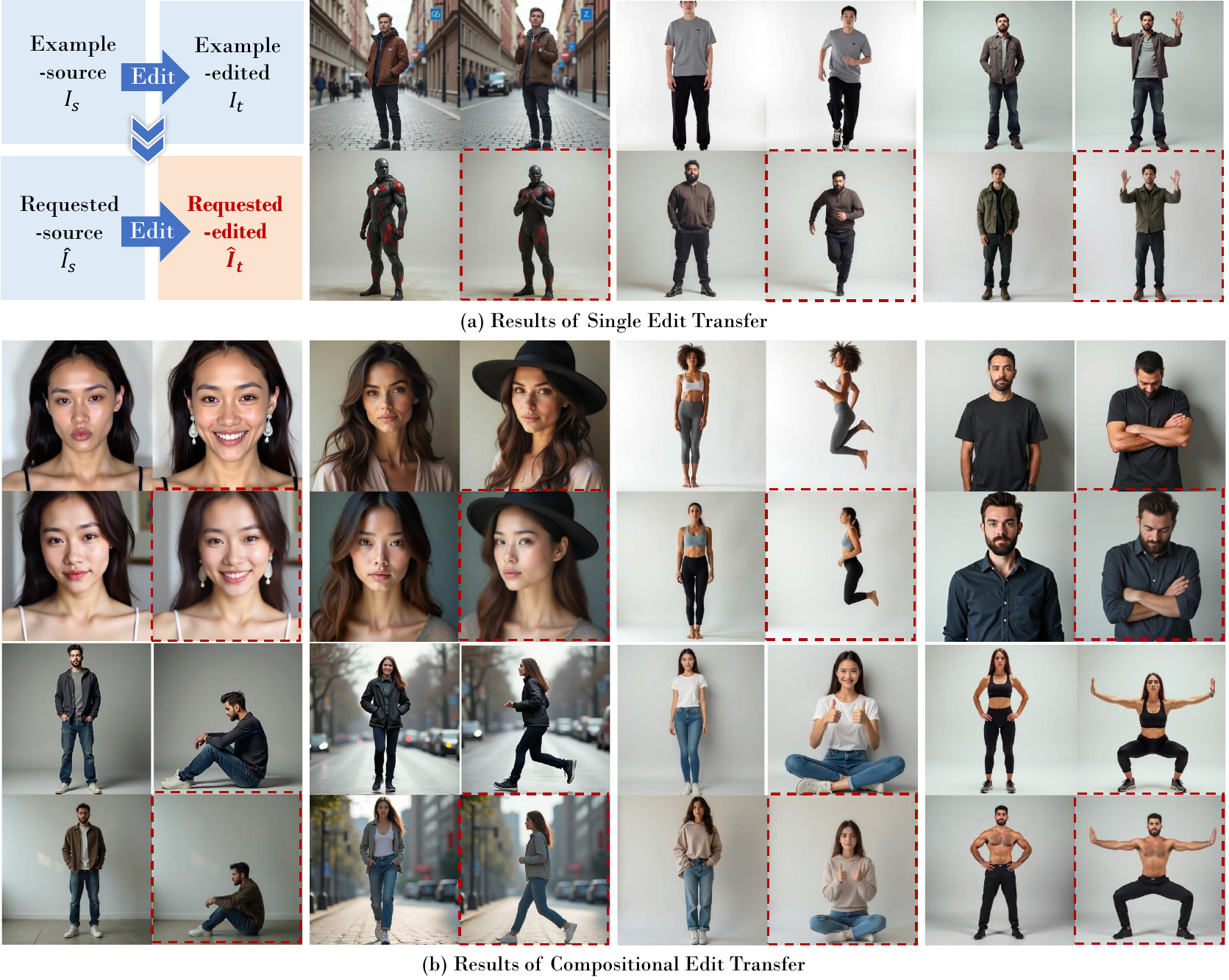}
    \vspace{-7 mm}  \caption{\textbf{\sys} exhibits impressive versatility to transfer visual exemplar pairs'edit into the requested source image, delivering high-quality (a) single-edit transformations as well as (b) effective compositional edits that seamlessly combine multiple modifications.
    }    \label{fig:results}
    \vspace{-2 mm}
\end{figure*}
\begin{figure*}[t]
    \includegraphics[width=\linewidth]{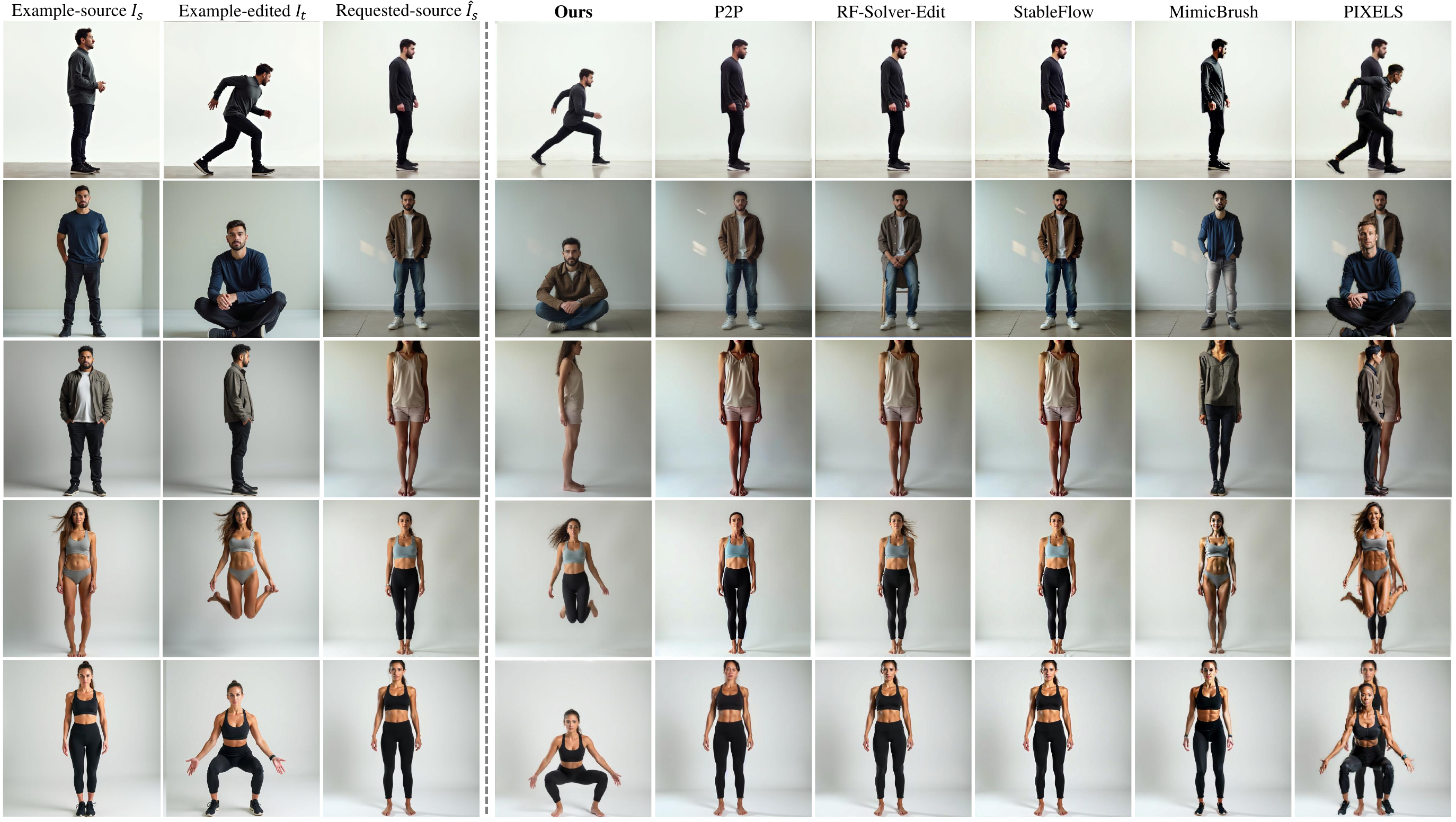}
    \vspace{-7 mm}
    \caption{\textbf{Qualitative comparisons.}
    Compared with TIE and RIE methods, our method consistently outperforms in various non-rigid editing tasks.
    We provide the detailed text prompt of TIE methods in ~\subsecref{IB}.
    }
    \label{fig:baseline}
    \vspace{-5 mm}
\end{figure*}

Unlike existing RIE methods that focus solely on style or appearance transfer, we introduce \sys, which learns the underlying transformation from a few-shot editing example and generalizes it to new images. 
As illustrated in \figref{framework}(a), an editing example is defined by a pair $(\mathcal{I}_s, \mathcal{I}_t)$—where $\mathcal{I}_s$ is the source image and $\mathcal{I}_t$ its edited result—and our goal is to learn the transformation from $\mathcal{I}_s$ to $\mathcal{I}_t$, then apply it to a new source image $\hat{\mathcal{I}}_s$ to produce the edited image $\hat{\mathcal{I}}_t$.
To this end, we first illustrate the rationale of the in-context capabilities of DiT-based T2I models in \subsecref{pre}, and then present our visual relation in-context learning paradigm for edit transfer in \subsecref{VR-IC}.

\subsection{Preliminary:DiT-based T2I Models}
\label{subsec:pre}                     
DiT-based T2I models (\eg, FLUX) adopt token-based representations and transformer architectures similar to LLMs, enabling in-context generation for recent T2I systems. 
In these models, images are tokenized into sequences $z\in\mathbb{R}^{N\times d}$ and processed alongside text tokens $c_T\in\mathbb{R}^{M\times d}$, with their spatial and sequential structures preserved across transformer layers.
In FLUX, each DiT block incorporates a MMA~\cite{pan2020multi} module after layer normalization~\cite{ba2016layer} to fuse image tokens $z$ and text tokens $c_T$.
Specifically, the concatenated sequence $[z; c_T]$ is projected into query ($Q$), key ($K$), and value ($V$) matrices, and MMA~\cite{pan2020multi} is computed as:

\begin{equation}
\vspace{-1 mm}
\text{MMA}([z; c_T]) = \text{softmax}\left(\frac{QK^\top}{\sqrt{d}}\right)V.
\label{attention1}  
\vspace{-0.5 mm}
\end{equation}
This bidirectional attention enables image tokens to interact both among themselves and with text tokens, forming the foundation of our framework.

\subsection{Visual Relation In-Context Learning}
\label{subsec:VR-IC}
Inspired by the success of few-shot in-context learning in LLMs, we extend this approach to the visual domain by proposing a visual relation in-context learning framework for \sys. 
Similar to LLMs, where carefully chosen in-context examples can greatly influence performance, we find that constructing high-quality examples is also crucial for our visual relation task.
Moreover, since T2I models have not encountered this specific \sys task during pre-training, fine-tuning becomes necessary to equip them with the requisite relational understanding.

\Paragraph{Constructing in-context samples.}
Unlike LLMs, which can process long token sequences, capturing relationships between separate images in FLUX requires a different approach. 
To encode a clear ``source image → edited image'' order, we arrange each example-request pair into a $2\times2$ composite: the first row contains the example pair $(\mathcal{I}_s, \mathcal{I}_t)$, and the second row contains the request pair $(\hat{\mathcal{I}}_s, \hat{\mathcal{I}}_t)$.
To ensure both pairs embody a consistent editing type, we synthesize multiple example pairs using the pre-trained FLUX.1-dev model and manually select those with visually aligned edits. 
We denote the resulting four-panel composite as a training sample $\mathcal{I}^\ast$ for a given editing type, as shown in \figref{framework}(a).
This design enables different sub-images to attend to each other via the MMA~\cite{pan2020multi} module, much like how text tokens interact in LLMs.
Notably, we use a simple text prompt to describe the interconnections among the four sub-images and to specify the editing type (\eg ``raising hands''), rather than providing a detailed description.
We construct $21$ distinct editing types for human, each represented by two diverse training examples, resulting in a dataset of 42 images.
More details are provided in the~\subsecref{dataset} and the effect of dataset size is discussed in \subsecref{AS}.
\Paragraph{Few-shot in-context fine-tuning.} 
Building on these carefully constructed four-panel samples, our next step is to leverage them for robust in-context fine-tuning.
As illustrated in ~\figref{framework}(b), building on the four-panel training structure, our objective is to generate the bottom-right image $\hat{\mathcal{I}}_t$ conditioned on 
 $\mathcal{I}_c = (\mathcal{I}_s, \mathcal{I}_t, \hat{\mathcal{I}}_s)$.
The input consists of conditional tokens $c_I$ (encoded from $\mathcal{I}_c$) and noisy tokens $z_0$ (encoded from $\hat{\mathcal{I}}_t$).
We add noise only to $z_0$ to obtain $z_t$, while keeping $c_I$ tokens clean.
Within each DiT block's MMA~\cite{pan2020multi} module, $z_t$ and $c_I$ are concatenated with $c_T$, and then projected into $Q$, $K$, and $V$. 
The MMA~\cite{pan2020multi} module then captures the relationships among these tokens as follows: 
\begin{equation}
\vspace{-1 mm}
\text{MMA}([z_t;c_T;c_I]) = \text{softmax}(\frac{QK^\top}{\sqrt{d}})V,
\label{attention}  
\end{equation}
where $[z_t;c_T;c_I]$ denotes the concatenation of noisy latent tokens, text tokens and conditional image token.
To further enhance in-context learning, we fine-tune a lightweight LoRA for the edit transfer task.
The conditional flow matching loss for fine-tuning is defined as:
\begin{equation}
\vspace{-1 mm}
\mathcal{L}_{CFM} = E_{t,p_t(z|\epsilon),p(\epsilon)}||v_\theta(z,t,c_T,c_I)-u_t(z|\epsilon)||^2,
\label{loss}  
\end{equation}
where $v_\theta(z,t,c_T,c_I)$ represents the velocity field parameterized by the model, $t$ is the timestep, and $u_t(z|\epsilon)$ is the target vector field conditioned on noise $\epsilon$.

\Paragraph{Edit transfer.} 
Empowered by the fine-tuned LoRA weights, the MMA~\cite{pan2020multi} module can now transfer the relational information from a test example pair to a new source image $\hat{\mathcal{I}}_s$.
Thus, \sys is cast as an image generation task: we initialize the bottom-right image token $z_T$ from random noise, concatenate it with $c_I$, and generate the output.
As demonstrated in~\figref{framework}(c), the model successfully generates the bottom right image $\hat{\mathcal{I}}_t$, effectively transferring the editing from the example pair to the new source image $\hat{\mathcal{I}}_s$ after denoising.

\begin{table}[!t]
\footnotesize
\renewcommand{\arraystretch}{1.2}
    \begin{tabular}{>{\centering\arraybackslash}p{1.4cm}@{\hspace{0.2cm}}>{\centering\arraybackslash}p{0.7cm}@{\hspace{0.4cm}}>{\centering\arraybackslash}p{1.2cm}@{\hspace{0.1cm}}>{\centering\arraybackslash}p{1.0cm}@{\hspace{0.1cm}}>{\centering\arraybackslash}p{1.0cm}@{\hspace{0.1cm}}>{\centering\arraybackslash}p{1.0cm}@{\hspace{0.1cm}}>{\centering\arraybackslash}p{0.6cm}}
    \toprule 
    \multirow{2}{*}{Method} & \multicolumn{3}{c}{TIE} & \multicolumn{2}{c}{RIE} & \multirow{2}{*}{\textbf{Ours}} \\
    \cmidrule(lr){2-4} \cmidrule(lr){5-6}
    & P2P~\cite{hertz2022prompt} & RFSE~\cite{wang2024taming} & 
    SF~\cite{avrahami2024stableflow}
    &
    MB~\cite{chen2024mimicbrush} &
    PX~\cite{biswas2025PIXELS}
    &\\
    \midrule
    CLIP-T $\uparrow$& 20.22 &19.83 & 18.67 & - & - &\textbf{22.58} \\
    CLIP-I $\uparrow$& -& - & - & 0.804 & 0.761 &\textbf{0.810} \\
    PickScore $\uparrow$&20.60   &20.96 & 20.91 &20.34  & 20.82 & \textbf{21.50} \\
    \bottomrule 
    \end{tabular}
    \vspace{-3 mm}
    \caption{\textbf{Quantitative comparisons of baselines and our \sys.} Bold indicates the best value.}
    \vspace{-3 mm}
    \label{tab:baseline}
\end{table}

\begin{figure}[!t]
    \includegraphics[width=\columnwidth]{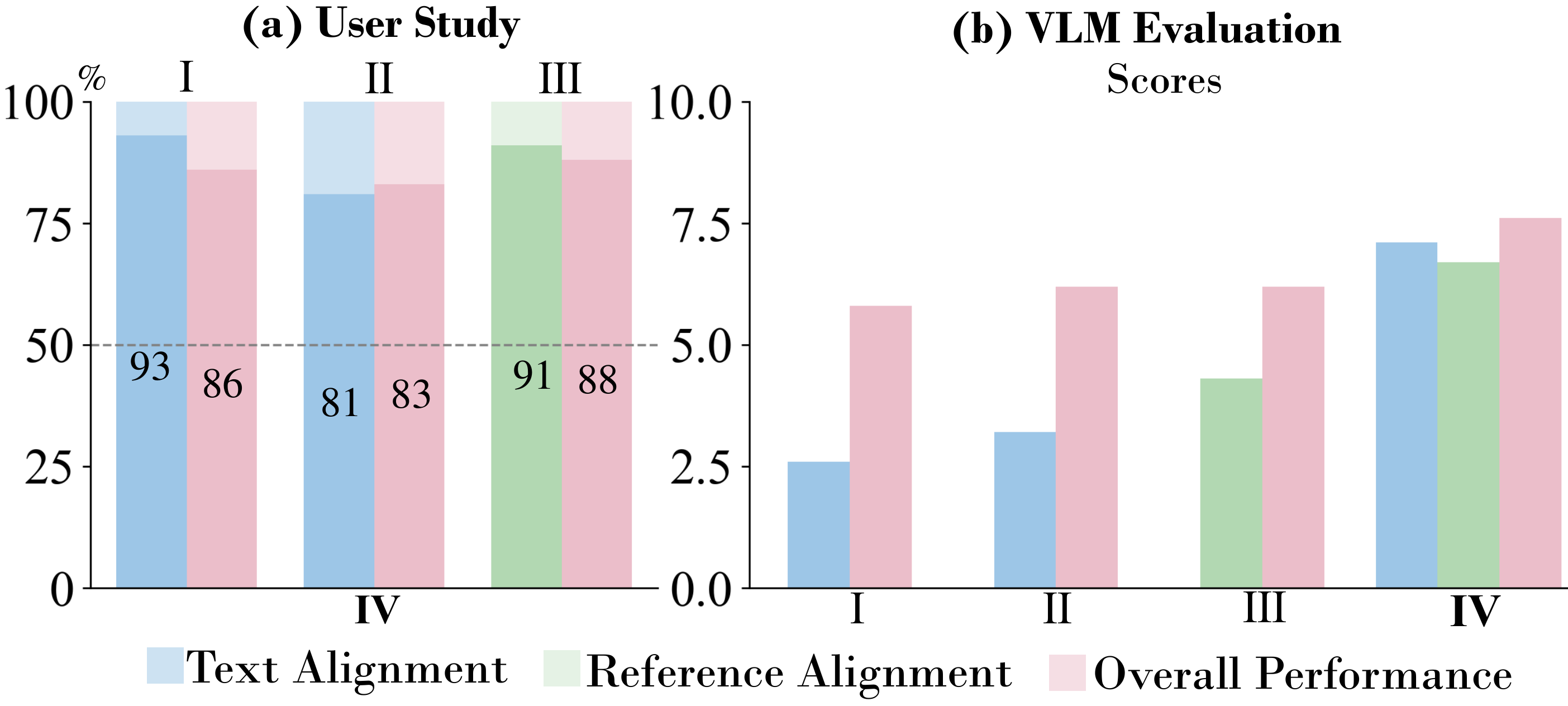}
    \vspace{-6 mm}
    \caption{\textbf{Results of user study and VLM evaluation.}
    We compare our \sys (\MakeUppercase{\romannumeral 4}) with P2P~\cite{hertz2022prompt} (\MakeUppercase{\romannumeral 1}), RF-Solver-Edit~\cite{wang2024taming} (\MakeUppercase{\romannumeral 2}) and MimicBrush~\cite{chen2024mimicbrush} (\MakeUppercase{\romannumeral 3}).
    (a)  The values show the proportion of users who prefer our method over the others.
    %
    %
    (b) The values represent the average scores given to each method by GPT-4o~\cite{openai2024gpt4technicalreport}.
    %
    %
    }
    \label{fig:user_study}
    \vspace{-6 mm}
\end{figure}

\section{Experiments}
\label{sec:Experiments}
\subsection{Training Setup}
We use FLUX.1-dev as the backbone model.
During training, we set the LoRA rank to $16$, the learning rate to $1e-4$, and the batch size to $4$.
The model is trained for $6000$ iterations on a single A100 (40GB) GPU using the Adam optimizer.
For more details, please refer to the~\subsecref{fine-tune}.

\subsection{Experimental Results}

\sys demonstrates impressive versatility by achieving both high-quality single-edit transfer and effective compositional edit transfer. 
As shown in \figref{results}(a), \sys consistently produces outstanding outputs for single-edit tasks: it accurately applies the transformation derived from example pairs while preserving crucial attributes, such as the identity, appearance, and background of the requested source image. 
Moreover, as illustrated in \figref{results}(b), \sys successfully performs compositional edit transfers in a single step, effectively combining multiple editing operations.

\begin{figure}[!t]
    \includegraphics[width=\columnwidth]{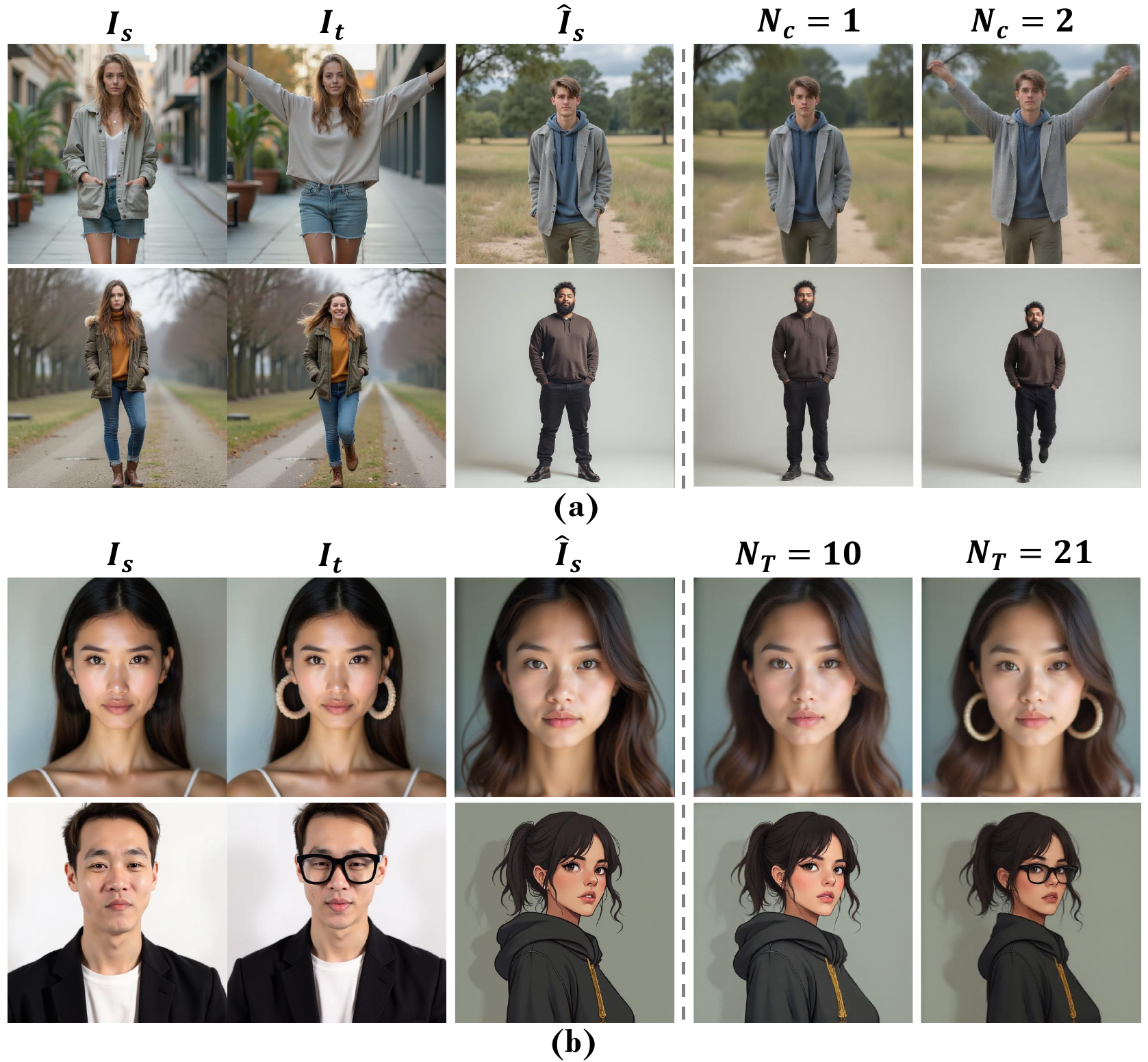}
    \vspace{-8 mm}
    \caption{\textbf{Influence of dataset scale.}
    (a)
    Setting the number of training samples per editing type to $N_c=2$ is sufficient for learning effective non-rigid edits, even when the total number of editing types $N_T=10$. 
    (b) Increasing $N_T=21$ further improves the model's ability to capture subtle local edits and enhances its generalization to cases where the editing example and query image are spatially misaligned.
    }
    \label{fig:ablation_number}
    \vspace{-6 mm}
\end{figure}

\subsection{Comparisons with Baselines}
\Paragraph{Baselines.}
We compare our method with existing representative TIE and RIE approaches.
\begin{compactitem}
\item \textbf{TIE methods}: 1) P2P~\cite{hertz2022prompt} is a classical text-based editing method;
2) RF-Solver-Edit (RFSE)~\cite{wang2024taming} and StableFlow (SF)~\cite{avrahami2024stableflow} propose new inversion methods for flow matching and adapt MasaCtrl~\cite{cao_2023_masactrl}'s principles into the DiT architecture, delivering enhanced editing capabilities.
\item \textbf{RIE method}: 
MimicBrush (MB)~\cite{chen2024mimicbrush} and PIXELS (PX)~\cite{biswas2025PIXELS} enable users to edit image regions by drawing on in-the-wild references without needing an exact source match.
\end{compactitem}
For TIE methods, we employ GPT-4o~\cite{openai2024gpt4technicalreport} to generate detailed text prompts followed by human revision, while for RIE methods—which support only a single reference image—we use $\mathcal{I}_{t}$ as the reference.

\Paragraph{Qualitative results.}
We present a visual comparison of
our approach against several baselines in \figref{baseline}.
It can be clearly observed that 
existing TIE methods struggle with complex spatial transformations.
For instance, in the fourth row, P2P~\cite{hertz2022prompt} fails to transfer into the expected pose.
While RF-Solver-Edit~\cite{wang2024taming} can perform simple non-rigid modifications, such as stretching arms, it fails to capture intricate leg movements required for a ``jumping" pose.
For RIE method, MimicBrush\cite{chen2024mimicbrush} focuses on transferring the texture from the reference image well, as evident in the copied clothing details in the second row of~\figref{baseline}.
However, it fails to transfer complex non-rigid transformations.
In contrast, our method successfully handles complex non-rigid edits, faithfully following the demonstrated examples.

\Paragraph{Quantitative results.}
We quantitatively evaluate our proposed method against baseline models using automatic metrics, human evaluations, and assessments by a vision-language model (VLM).
For further implementation details, please refer to the~\subsecref{ED}.

\emph{Automatic Metrics.} 
To evaluate TIE methods, we employ CLIP-T~\cite{radford2021learning} to measure the alignment between text and image.
For RIE method, we utilize CLIP-I~\cite{radford2021learning} to quantify the similarity between the edited image $\hat{\mathcal{I}}_t$ and the reference image $\mathcal{I}_t$.
Furthermore, we incorporate PickScore~\cite{Kirstainpickscore} to assess the overall quality of the generated images.
As presented in \tabref{baseline}, our \sys demonstrates strong alignment with both text prompts and reference images while also achieving superior overall image quality.

\emph{User Study.} 
We conduct a user preference study using pairwise comparisons.
For text-based image editing (TIE), we evaluate text alignment to assess how well the generated images correspond to the provided descriptions.
For reference-based image editing (RIE), we measure reference alignment, determining how accurately the generated images reflect the intended transformations based on the reference image.
Additionally, we assess overall user preference to gauge perceived quality.
As shown in \figref{user_study} (a), our method achieves an outstanding preference rate exceeding $80\%$ across all aspects, consistently outperforming the baselines in all three evaluation criteria.
\begin{figure}[t]
    \includegraphics[width=\columnwidth]{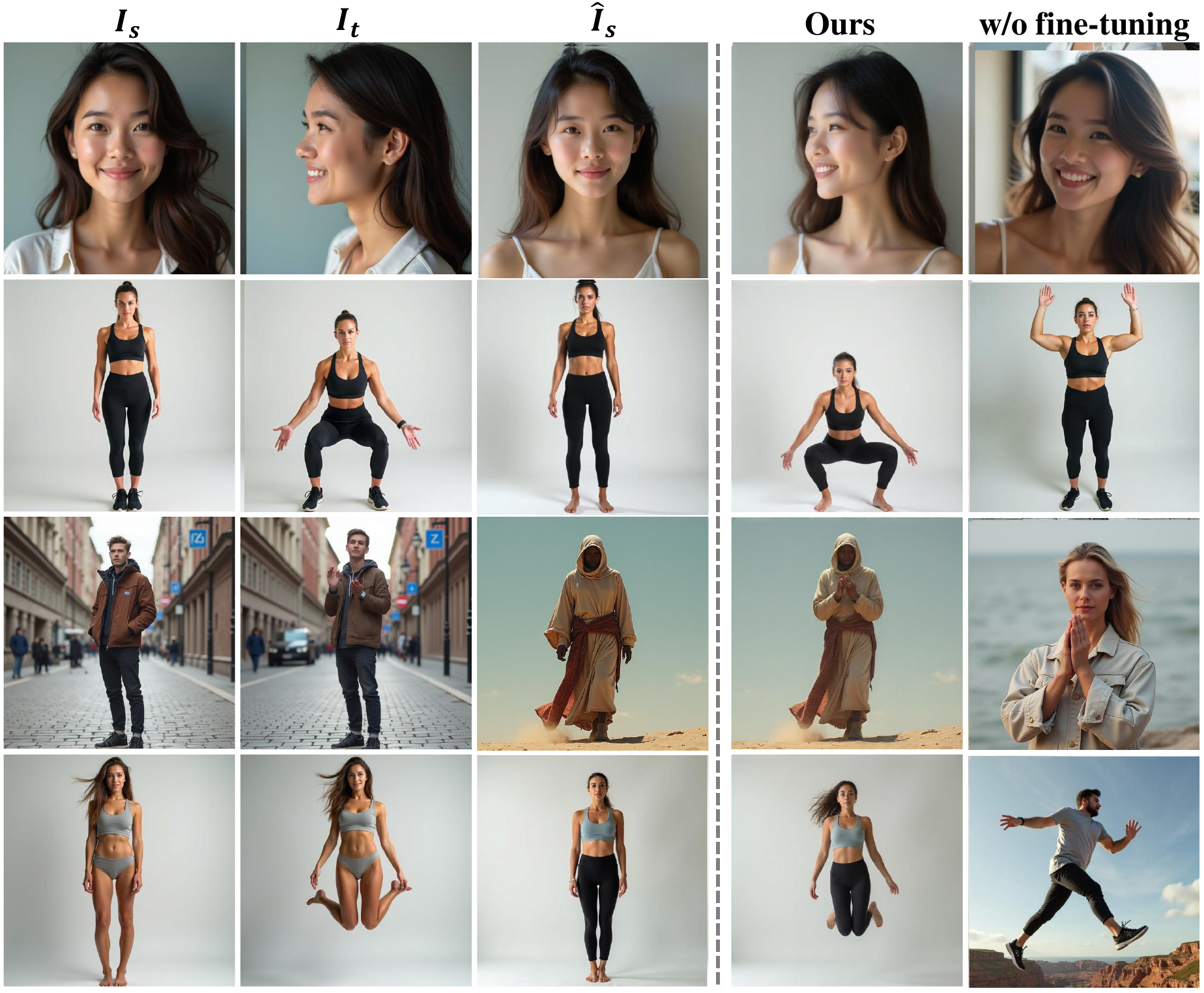}
    \vspace{-7 mm}
    \caption{\textbf{Ours \vs w/o fine-tuning.}
    Without fine-tuning, Flux can only capture some of the pose information identified in $I_t$ regardless of the relation between $I_s \to I_t$ and  $\hat{I}_s$. 
     In contrast, with our few-shot fine-tuning, the model effectively learns the visual relation from example pairs and applied to $\hat{I}_s$.
     }
    \label{fig:flux}
    \vspace{-3 mm}
\end{figure}

\emph{VLM Evaluation.}
We adopt the scoring rules based on the LMM Score methodology~\cite{huang2025diffusion}
to evaluate text alignment, reference alignment, and overall performance via VLM GPT-4o~\cite{openai2024gpt4technicalreport}.
Each criterion is scored on a scale from $1$ to $10$, with higher scores indicating better performance.
As shown in \tabref{baseline}, our model achieves the highest scores against other baselines across all criterias.


\begin{figure}[t]
    \includegraphics[width=\columnwidth]{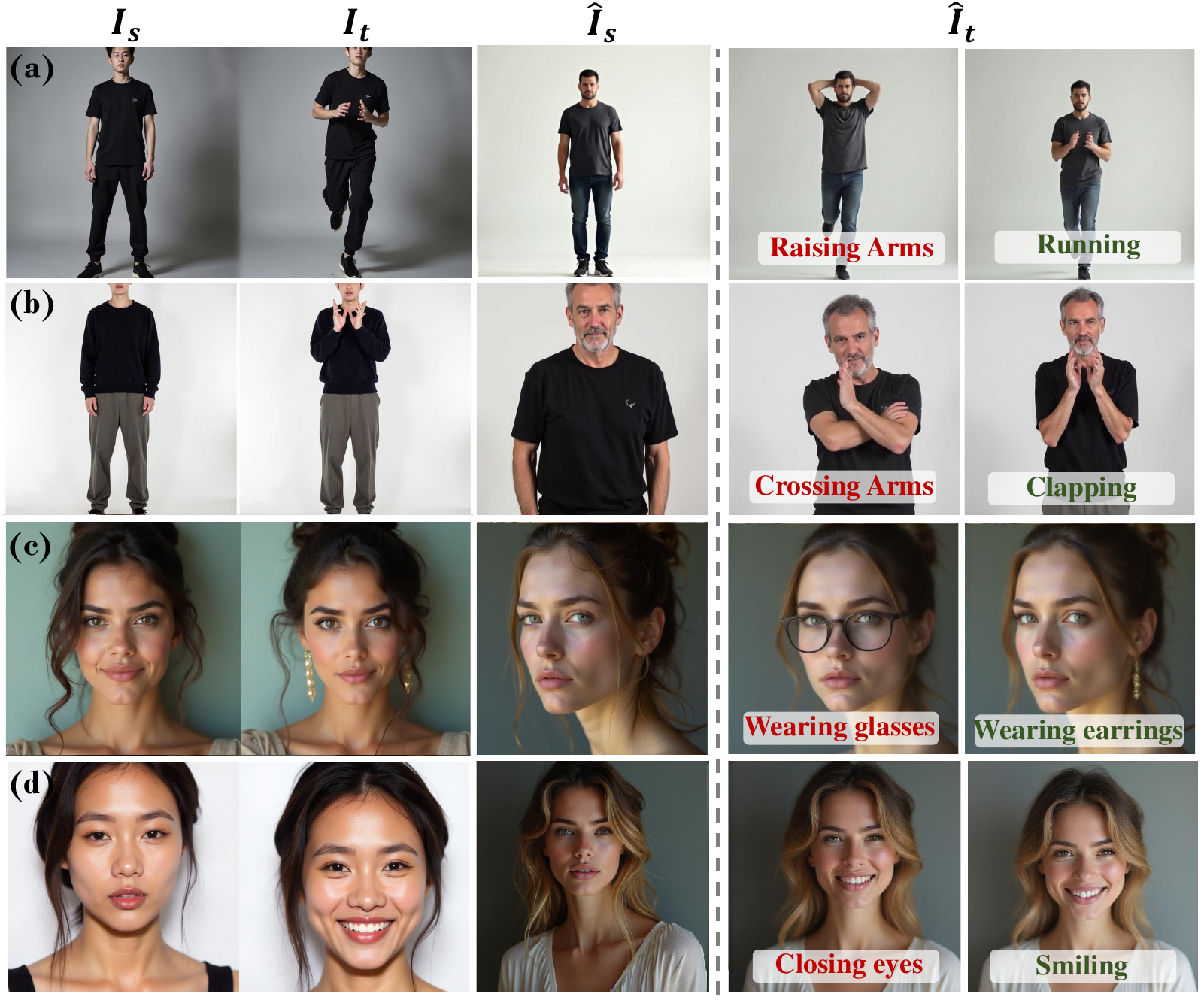}
    \vspace{-7 mm}
    \caption{\textbf{Investigating the alignment between text and visual example pairs.}
    When the text prompt and visual demonstrations convey different semantics, the generated images $\hat{\mathcal{I}}_t$ tend to (a)(b) exhibit mixed semantics from both sources, and either (c) follow the text or (d) the visual demonstrations.
    Note that the \textcolor{color1}{red} label indicates misalignment, while \textcolor{color2}{green} label indicates alignment.
    }
    \label{fig:ablation_prompt}
    \vspace{-6 mm}
\end{figure}
\begin{figure*}[t]
    \includegraphics[width=\textwidth]{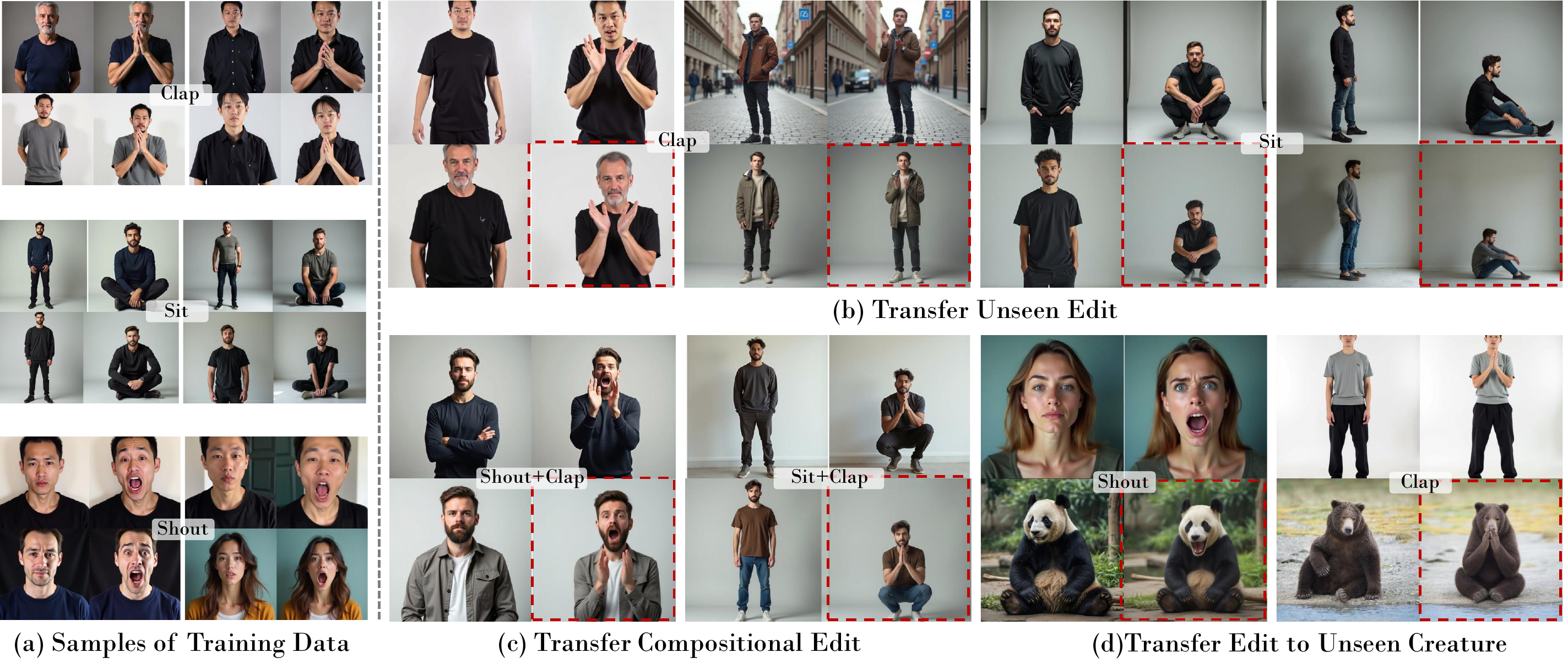}
    \vspace{-7 mm}
    \caption{\textbf{Generalization performance of \sys.}
Our model demonstrates remarkable generalization by: (b) Generating novel pose variations within a given editing type, even if such variations were unseen during training; (c) Flexibly combining different editing types; (d) Transferring its capabilities across other species.
    }
    \label{fig:generalization}
    \vspace{-6 mm}
\end{figure*}
\subsection{Ablation Study}
\label{subsec:AS}
We perform an ablation study to investigate how in-context sample size and fine-tuning influence our visual relation in-context learning approach.

\Paragraph{Influence of dataset scale.}
Effective in-context examples are crucial for \sys. 
To assess dataset scale, we vary the number of editing types ($N_T$) and examples per type ($N_c$). 
With $N_T=10$, training with $N_c=1$ fails to produce the desired ``raising arms'' transformation, whereas $N_c=2$ succeeds (see \figref{ablation_number}(a)).
Keeping $N_c=2$ fixed, increasing $N_T$ from $10$ to $21$ not only captures major non-rigid transformations but also improves fine-grained detail transfer (\eg, ``wearing earrings,'' as shown in \figref{ablation_number}(b)), and enhances adaptation to spatial misalignment between visual example pairs and query images. 
We hypothesize that, similar to the benefits of multi-task fine-tuning~\cite{liumft} observed in LLMs, increasing editing types diversity enhances the model’s in-context learning ability. 
Thus, we adopt $N_T=21$ and $N_c=2$ in our final configuration.

\Paragraph{Ours \vs w/o fine-tuning.}
To validate the efficiency of our in-context fine-tuning strategy, we compare our model with FLUX.1-dev without fine-tuning.
As shown in~\figref{flux}, 
although the model can capture some semantic aspects of the pose identified in $I_t$, it fails to understand the detailed relationships from $I_s \to I_t$ and the fine-grained information of $I_s$.
For instance, while the image in the last row exhibits a ``jumping'' motion, the resulting posture significantly deviates from that in $\mathcal{I}_t$.
In contrast, our fine-tuned model successfully transfers the edits demonstrated in the example pairs to the requested image $\hat{\mathcal{I}}_s$, verifying that even a small set of in-context examples enables the model to learn precise visual relationships.
%

\begin{figure}[t]
    \includegraphics[width=\columnwidth]{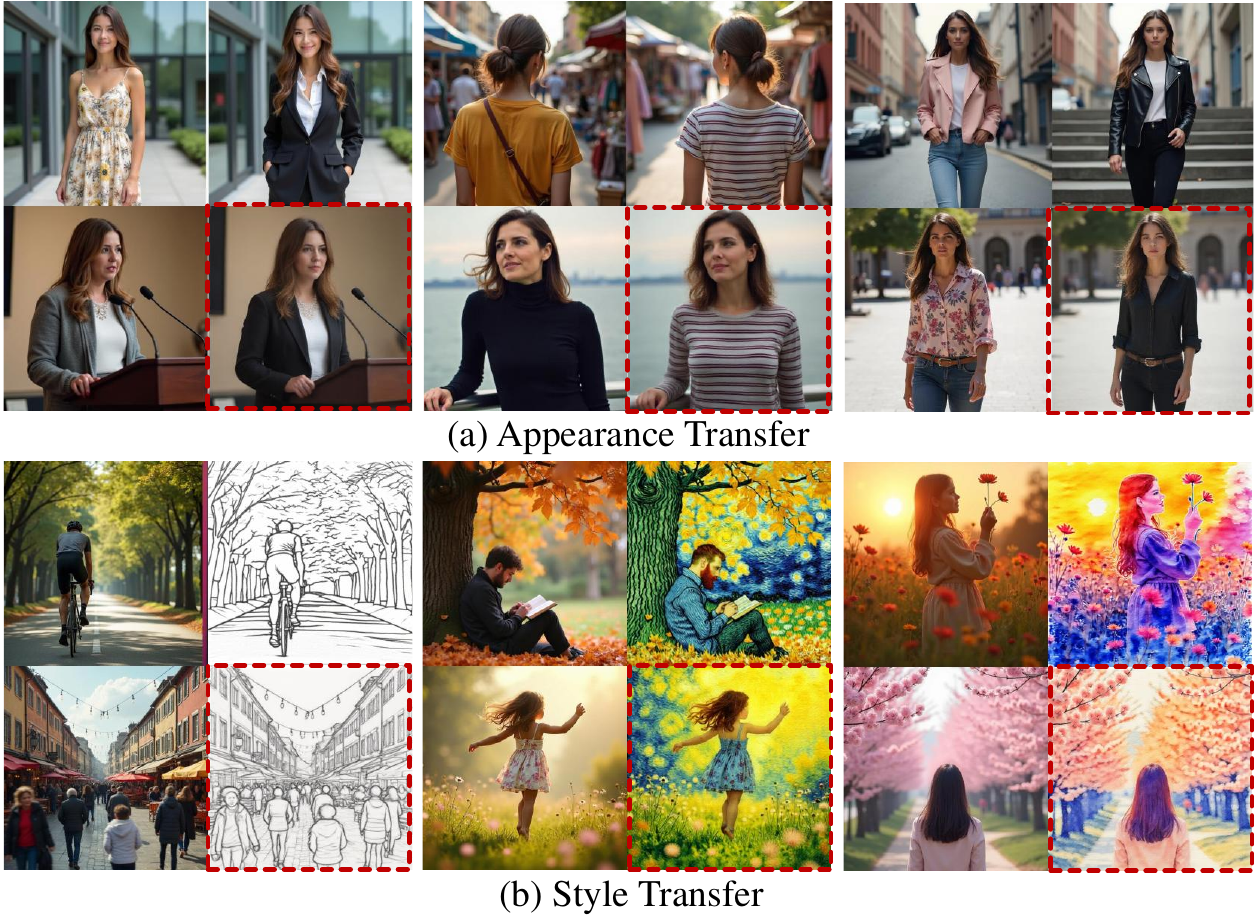}
    \vspace{-8 mm}
    \caption{\textbf{Applications to appearance and style transfer.}
    Our proposed method can also be applied to other editing types such as appearance and style transfer.}
    \label{fig:other}
    \vspace{-3 mm}
\end{figure}
\subsection{Discussion}
\Paragraph{Alignment between text and visual example pairs.}
In our framework, both text and visual example pairs guide the editing process. 
To evaluate their influence, we deliberately introduce a semantic mismatch between them. 
As shown in \figref{ablation_prompt}, the output may (1) blend both cues of text and visual guidance, (2) follow the text prompt, or (3) follow the visual example pairs. 
For instance, in \figref{ablation_prompt}(a), the leg pose in $\hat{\mathcal{I}}_t$ closely mirrors the visual demonstrations, while the arm pose follows the ``raise'' prompt. 
In \figref{ablation_prompt}(c) and \figref{ablation_prompt}(d), the output tends to follow the text prompt or visual demonstrations.
These results underscore the need to align text and visual guidance for consistent editing transfer.

\Paragraph{The generalization capability.}
Our model exhibits remarkable generalization.
1) It can transfer new pose styles for a training editing type, which has never been seen during training, such as different sitting styles in ~\figref{generalization}(b); 
 2) it can flexibly combine any poses of editing type, as illustrated in~\figref{generalization}(c), when the source-target example pairs include both ``shouting'' and ``clapping'' poses, the model successfully transfers both; 
 3) and it can even transfer edits to a different creature (\eg panda) that is not present in the training data, as shown in~\figref{generalization}(d).

\Paragraph{Application to other editing types.}
While our focus is on non-rigid editing tasks, the proposed EditTransfer can also be extended to other editing types such as appearance and style transfer, as demonstrated in~\figref{other}.


\section{Conclusions and Future Work}
\label{sec:Conclusion}
In this work, we introduce the novel \sys setting, where a transformation is learned directly from only a single source–target editing pair and then applied to a new image.  
Departing from text-centric and appearance-focused paradigms, we demonstrate that a DiT-based T2I model, equipped with visual relation in-context learning and lightweight LoRA fine-tuning, can capture complex spatial edits—even with only a handful of examples. 
Our results highlight the surprising potential of minimal data to yield sophisticated, non-rigid transformations once thought beyond the scope of conventional TIE and RIE approaches.

Looking ahead, \sys opens up promising directions: by gathering a small set of examples spanning various editing types, creators can easily achieve a wide range of visual effects.  
In future work, we aim to explore more advanced editing tasks and broaden the applicability of non-rigid editing to diverse species and scenarios. 
{
    \small
    \bibliographystyle{unsrt}
    \bibliography{main}
}
\clearpage
\appendix
\appendixpage

\tcbset{
  aibox/.style={
    top=10pt,
    colback=white,
    enhanced,
    center,
  }
}
\newtcolorbox{AIbox}[2][]{aibox, title=#2,#1}

\section{Implementation Details}
\subsection{Dataset}
\label{subsec:dataset}
\begin{figure}[!t]
\centering{\includegraphics[width=\columnwidth]{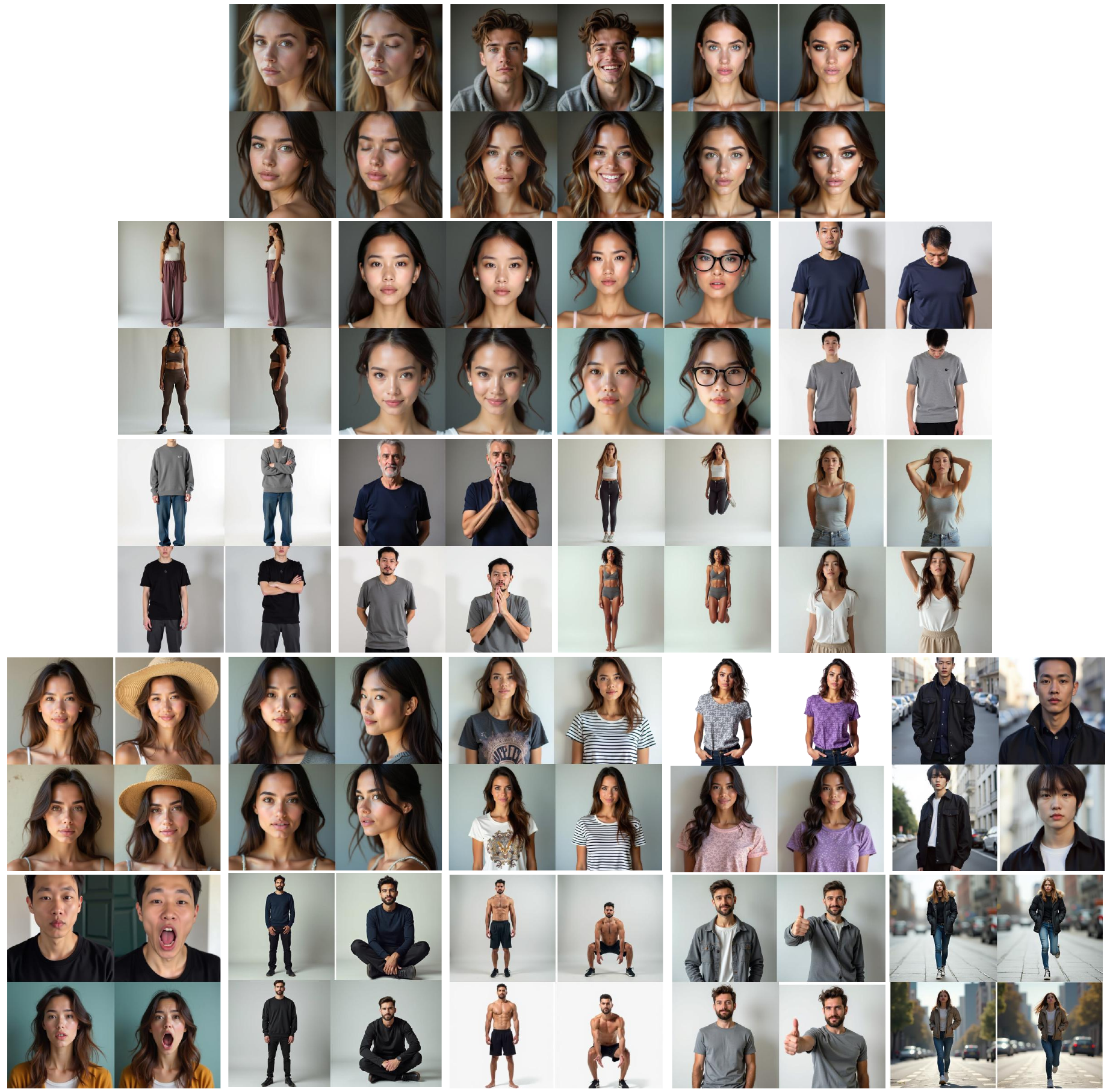}}
    \vspace{-6 mm}
    \caption{\textbf{Image samples of each editing type in the dataset.}
    }
    \label{fig:dataset}
    \vspace{-6 mm}
\end{figure}

Leveraging its inherent in-context generation capability, we utilize  FLUX.1-dev\footnote{\url{https://github.com/black-forest-labs/flux}} to generate diverse two-panel images, each consisting of a source image $\mathcal{I}_s$ and a corresponding ``edited" image $\mathcal{I}_t$.
We carefully select the generated pairs to ensure subject identity consistency, that each editing pair demonstrates only a single transformation, and that the source image $\mathcal{I}_s$ is in the front view.
We manually select two pairs of the same editing type that are closely aligned in terms of viewpoint, scale, and transformation style.
In total, our dataset comprises $42$ images spanning $21$ editing types.
\figref{dataset} presents one sample from each editing type in the dataset.
Each four-panel image is accompanied by a text prompt that describes the relationships between the grids and roughly indicates the motion.
\figref{prompt} shows two prompt examples for single and compositional edit transfer.

\subsection{Fine-tuning and Inference}
\label{subsec:fine-tune}
We modify the LoRA training codes building on FLUX from the AI-Toolkit repository\footnote{\url{https://github.com/ostris/ai-toolkit}} to align with our proposed framework.
The fine-tuning process takes approximately $16$ hours for $6000$ iterations on a single A100 (40GB).
For inference, we modify the denoising process of FLUXInpaintPipeline\footnote{\url{https://github.com/huggingface/diffusers/blob/main/src/diffusers/pipelines/flux/pipeline_flux_inpaint.py}}.
The number of denoising steps is set to $35$, and the guidance scale is set to $1$.

\section{Details of Comparisons with Baselines }

\subsection{Implementation Details of Baselines}
\label{subsec:IB}
We use the official code for P2P\footnote{\url{https://github.com/google/prompt-to-prompt}}, RF-Solver-Edit\footnote{\url{https://github.com/wangjiangshan0725/RF-Solver-Edit}}, and MimicBrush\footnote{\url{https://github.com/ali-vilab/MimicBrush}}.
The masks used in MimicBrush~\cite{chen2024mimicbrush} are  obtained by SAM\footnote{\url{https://github.com/facebookresearch/segment-anything}}.
For P2P and RF-Solver-Edit~\cite{wang2024taming}, we generate prompts describing $\hat{\mathcal{I}}_s$ and $\hat{\mathcal{I}}_t$ using GPT-4o~\cite{openai2024gpt4technicalreport}, with human revisions.
The target prompts used by the TIE methods in~\figref{baseline} of the paper are presented in \figref{baseline_prompt}.

\subsection{Evaluation Details}
\label{subsec:ED}
We utilize the CLIP-T Score, CLIP-I Score based on the CLIP ViT-L/14 model, and the Pickup Score, as implemented in\footnote{\url{https://github.com/showlab/loveu-tgve-2023/blob/main/scripts/run_eval.py}} as our evaluation metrics. 
For VLM Score, we follow the procedure of LMM Score~\cite{huang2025diffusion}.
The evaluation process are shown in~\figref{vlm_score}, t
We use GPT-4o~\cite{openai2024gpt4technicalreport} to complete VLM Score.

\subsection{Details of User Study}
The user study comprises 198 tasks. 
In each task, participants are shown a source image along with two edited images: one generated by our proposed method and the other by a randomly selected baseline method.
The order of the images is shuffled to ensure unbiased evaluation.
A simplified text prompt is provided for comparison with P2P~\cite{hertz2022prompt} and RF-Solver-Edit~\cite{wang2024taming}, while the reference image is provided for comparison with MimicBrush~\cite{chen2024mimicbrush}.
There are three questions for the participants to answer: 
\begin{compactitem} 
\item \textbf{Text-Alignment (ours \vs text-based methods):} Which image aligns better with the $[$target prompt$]$''? 
\item \textbf{Reference-Alignment (ours \vs reference-based methods):} Which image contains subjects that are more aligned with the  $[$reference image$]$ in terms of motion (or decoration)?
\item \textbf{Overall Performance:} Which image do you prefer overall? 
\end{compactitem}

\section{Additional Results}
\begin{figure*}[!h]
    \includegraphics[width=\linewidth]{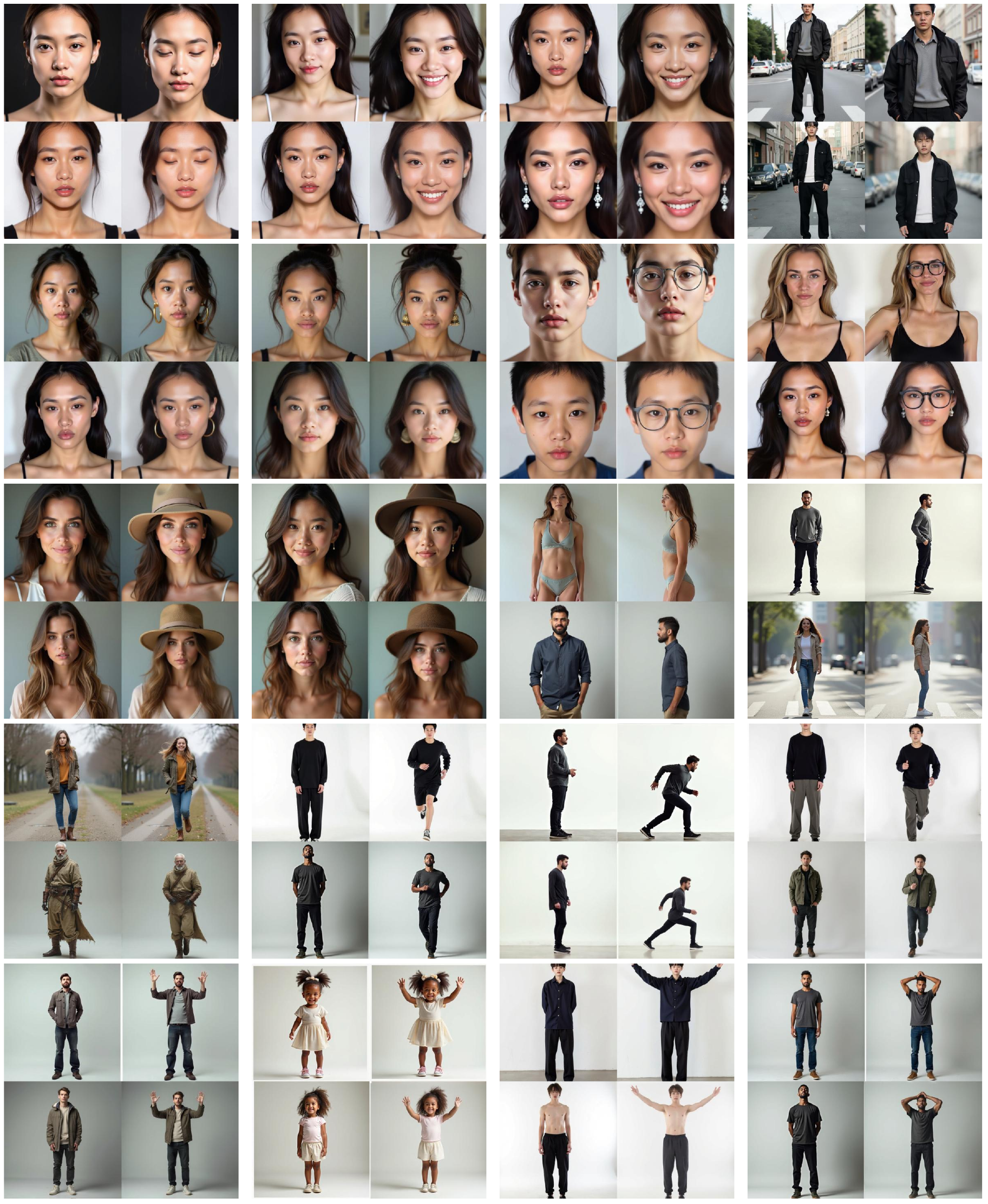}
    \vspace{-6 mm}
    \caption{\textbf{Additional experimental results of single edit transfer.}}
    \label{fig:editing1}
    \vspace{-3 mm}
\end{figure*}

\begin{figure*}[!h]
    \includegraphics[width=\linewidth]{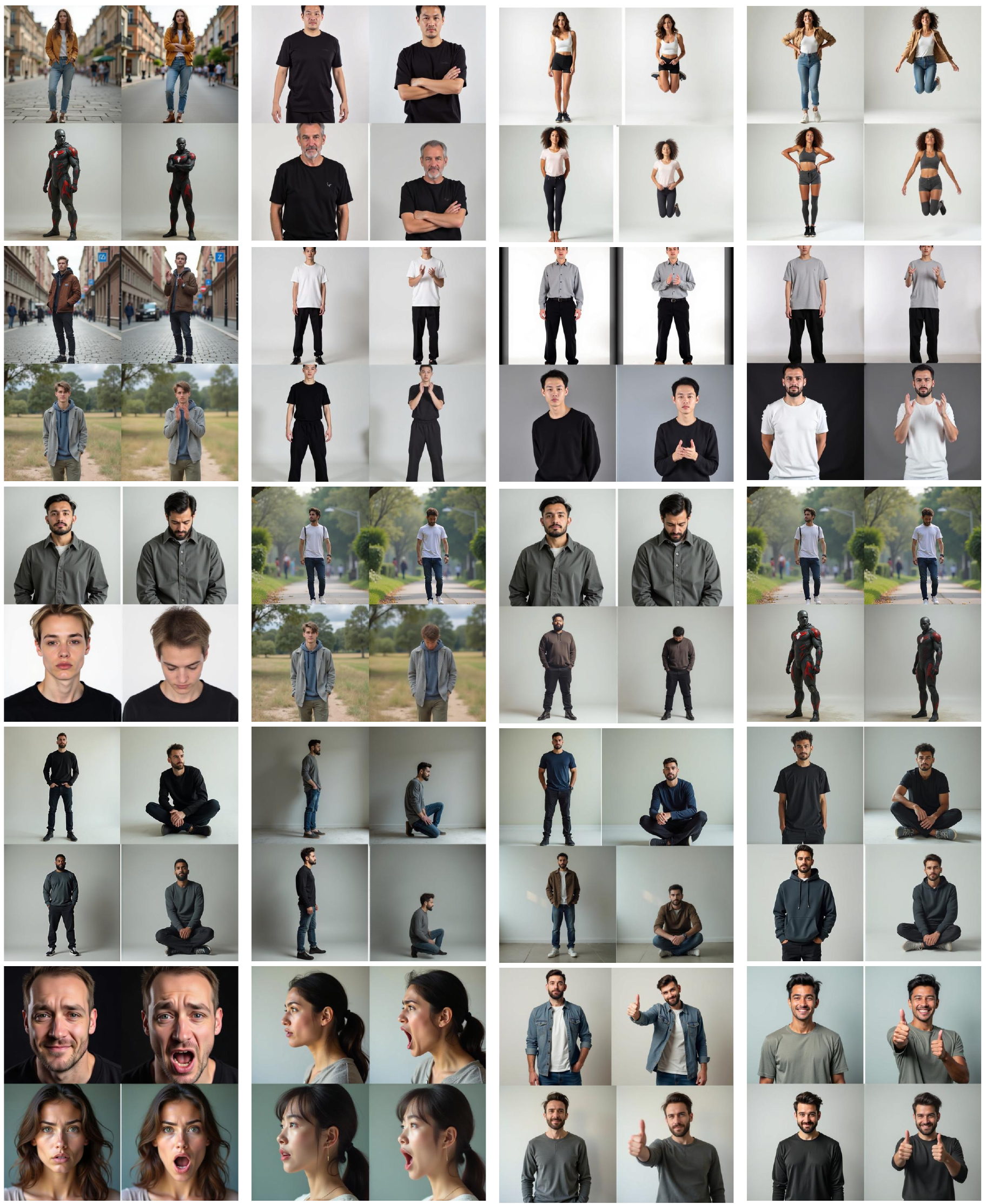}
    \vspace{-6 mm}
    \caption{\textbf{Additional experimental results of single edit transfer.}}
    \label{fig:editing2}
    \vspace{-3 mm}
\end{figure*}

As shown in~\figref{editing1} and~\figref{editing2}, our method performs well in handling complex spatial transformations.
Additionally, our method generalizes effectively to compositional edit transfer, as demonstrated in~\figref{editing2}.

\begin{figure*}[t]
    \begin{minipage}[t]{0.48\textwidth}
        \centering
        \begin{AIbox}{Text Prompts}
        This is a four-panel image grid: [TOP-LEFT]: The original source image. [TOP-RIGHT]: An edited version of the [TOP-LEFT] image, transformed to depict a running motion. [BOTTOM-LEFT]: A second original source image. [BOTTOM-RIGHT]: An edited version of the [BOTTOM-LEFT] image, applying the same running motion transformation as used in [TOP-RIGHT].
        \end{AIbox} 
        \vspace{-5 mm}
        \caption{\textbf{Prompt example edit transfer.}}
        \label{fig:prompt}
    \end{minipage}
    \hfill
    \begin{minipage}[t]{0.48\textwidth}
        \centering
        \begin{AIbox}{Text Prompts}
        {\color{deepblue}\bf Row 1:} a man lowers his head, facing the camera \\
        {\color{deepblue}\bf Row 2:} a sitting man with legs crossed and hands resting on his lap \\
        {\color{deepblue}\bf Row 3:} a standing woman turning right \\
        {\color{deepblue}\bf Row 4:} a jumping woman with arms extended sideways and legs bent at the knees \\
        {\color{deepblue}\bf Row 5:} a woman squats while clasping her hands together
        \end{AIbox} 
        \vspace{-5 mm}
    \caption{\textbf{Text prompts generated by GPT-4o~\cite{openai2024gpt4technicalreport} for TIE methods.}}
        \label{fig:baseline_prompt}
    \end{minipage}
\end{figure*}

\begin{figure*}[!t] 
\vspace{-1em}
\begin{AIbox}{Prompt Template}
{\color{black}\bf Step 1: Role Play} \\
{\color{deepblue}\bf User:}
I invite you to participate in an experiment on 'Evaluation of Image Editing Performance with Different Methods'. Be patient and focus. I will pay you tips according to your performance. Are you ready now? \\
{\color{Brown}\bf Assistant:}
I'm ready and intrigued! Let's dive into the experiment. How can I assist you with the evaluation?\\
{\color{deepblue}\bf User:}
Step 1: I will provide a source image, and an editing prompt that describes the target edited image. Please first view the images and describe the source image. Then describe the editing process in simple words. \\
{\color{Brown}\bf Assistant:}
Please go ahead and provide the source image, task indicator, and editing instruction. I'll start by describing the source image once I see it. \\
{\color{deepblue}\bf User:}
** Source Image: **[Image]\\
** Prompt:** [Prompt]
\\
{\color{Brown}\bf Assistant:}
[Answer]
\tcblower
{\color{black}\bf Step 2: Task Definition} \\
{\color{deepblue}\bf User:}
Step 2: Now there are 3 edited images which are edited from the source image by 3 different methods with the same one editing prompt mentioned before. You are required to first pay enough careful examination of the 3 edited images. Then you need to score each one based on each of the four following evaluation factors separately, resulting in 2 sub-scores ranging from 1 to 10, respectively. A higher score means better. \\
The two factors are:\\
  1. **Editing Accuracy**: Evaluate how closely the edited image (I2) adheres to the specified editing prompt (P), measuring the precision of the editing.  \\
  - Note that details described in P, such as motion and decorations, are important factors to consider when rating.  
  \\      
  2. **Overall Performance**: Rate the overall quality of the edited image (I2) in terms of coherence and realism. \\ 
Before scoring, carefully specify what to be concerned in each factor in this case. This can help you score more accurately. After that, I will upload the 3 edited images one by one. When I upload one image each time, you first give a detailed analysis based on the factors. \\
After I upload each image, you can score them in JSON format as shown:\\
{\{ ``Editing Accuracy": 2}, ``{Overall Performance": 3 \}}
\\
{\color{Brown}\bf Assistant:}
[Answer] 
\tcbline
{\color{black}\bf Step 3: Scoring Process} \\
{\color{deepblue}\bf User:}
[Image1] \\
{\color{Brown}\bf Assistant:}
[Answer1]\\
{\color{deepblue}\bf User:}
[Image2] \\
{\color{Brown}\bf Assistant:}
[Answer2] \\
{\color{deepblue}\bf User:}
[Image3]  \\
{\color{Brown}\bf Assistant:}
[Answer3]
\end{AIbox} 
\vspace{-1em}
\caption{\textbf{Prompt template of VLM score.}}
\label{fig:vlm_score}
\vspace{-1em}
\end{figure*}

\begin{figure*}[!h]
    \includegraphics[width=\linewidth]{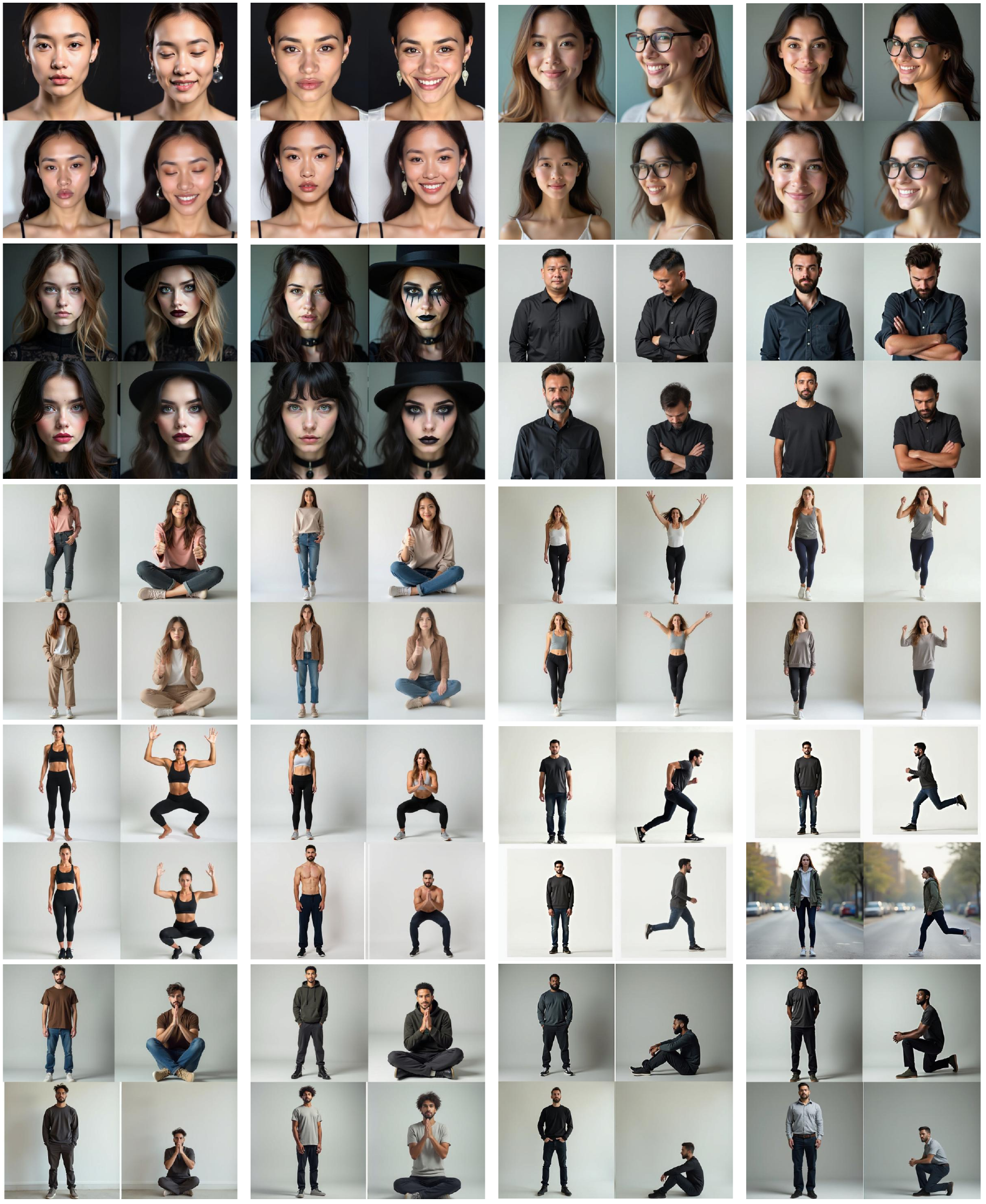}
    \vspace{-6 mm}
    \caption{\textbf{Additional experimental results of compositional edit transfer.}}
    \label{fig:editing3}
    \vspace{-3 mm}
\end{figure*}

\end{document}